\documentclass{article}

\usepackage[preprint]{neurips_2024}
\setcitestyle{numbers,square,comma}
\usepackage{graphicx}
\usepackage{amsmath}
\usepackage{url}
\usepackage{booktabs}
\usepackage{caption}

\usepackage{titlesec}
\usepackage{amssymb}
\usepackage{subcaption}
\usepackage{adjustbox}
\usepackage{array}
\usepackage{makecell} 
\usepackage{rotating} 
\usepackage{dsfont} 
\usepackage{float}
\usepackage{tikz}
\usepackage{pgfplots}
\pgfplotsset{compat=1.18}
\usetikzlibrary{pgfplots.groupplots}
\usepackage{newtxtext}
\usepackage[font=small,labelfont=bf]{caption}

\titleformat{\section}
  {\fontsize{12}{14}\bfseries\rmfamily\selectfont}
  {\thesection}{0.5em}{}
\titleformat{\subsection}
  {\fontsize{11}{13}\bfseries\rmfamily\selectfont}
  {\thesubsection}{0.5em}{}
\titlespacing{\section}{0pt}{*2}{*1.5}
\titlespacing{\subsection}{0pt}{*1.5}{*1}

\title{Multiscale Video Transformers for Class Agnostic Segmentation in Autonomous Driving}

\author{%
  Leila Cheshmi\thanks{Corresponding author. Email: leila.cheshmi@ontariotechu.net} 
  \And
  Mennatullah Siam
}

\date{} 

\begin{document}

\maketitle 

\begin{abstract}

\textit{Ensuring safety in autonomous driving is a complex challenge that requires handling unknown objects and unforeseen driving scenarios. In this work, we focus on developing multiscale video transformers that are capable of detecting unknown objects using only motion cues. Video semantic/panoptic segmentation often relies on a closed set of known classes seen during training, overlooking unknown or novel categories. Recent visual grounding with multimodal large language models (LLMs) is computationally expensive, as they utilize the full power of LLMs, especially when aiming for higher accuracy using larger variants and when considering pixel-level output. We envision an interplay between such models and light-weight efficient models designed for safety purposes, such as ours. To address this limitation in real-time safety-critical scenarios, the video class-agnostic segmentation (VCAS) task has been introduced, which allows for identifying such unseen objects. We propose an efficient video transformer that is trained end-to-end for video class agnostic segmentation without using optical flow as input. Our method relies on a novel multi-stage multiscale query-memory decoding and a scale-specific random drop-token to ensure efficiency and accuracy. It maintains detailed spatiotemporal features across decoding layers using a shared, learnable memory module. Unlike conventional query-based multiscale decoders that compress features—risking loss of fine spatial details—our memory-centric design preserves high-resolution information at multiple scales, improving segmentation quality. Moreover, our approach avoids computational overhead from requiring optical flow, supporting real-time performance. We evaluate our method on multiple video segmentation benchmarks, including generic ones such as DAVIS’16 and urban driving scenes, including KITTI and Cityscapes. Our method consistently outperforms traditional multiscale baselines with a considerable margin, while being efficient in GPU memory and run-time. Our results highlight a promising direction for real-time, robust dense prediction using video transformers in safety-critical robotics applications.}

\end{abstract}

\section{Introduction}

Video segmentation \cite{9966836,10.1145/3391743} is a fundamental yet challenging task in computer vision that involves identifying objects with specific semantic or physical properties across video frames. It plays a vital role in autonomous driving, where understanding dynamic scenes is essential for safe navigation. Video segmentation methods are typically categorized based on the level of human involvement during inference: interactive, semi-automatic, and automatic. Automatic video object segmentation (AVOS) approaches include optical flow-based two-stream models~\cite{zhou2020matnet,ren2021reciprocal,siam2019video} and more recent transformer-based fusion methods~\cite{zhou2020matnet,ren2021reciprocal,yuan2023isomer}. Semantic \cite{long2015fully}, instance \cite{he2017mask} and panoptic \cite{kirillov2019panoptic} segmentation and their extension to videos \cite{yang2019video,kim2020video} have been heavily explored. One of the most prominent methods, Mask2Former \cite{cheng2022masked}, introduced a universal segmentation technique that relies on a mask classification loss and a transformer decoder using masked attention and learned multiscale representations. 

However, many of these models are trained on datasets with a closed-set of known object classes. This leads to a major limitation in safety-critical applications such as autonomous driving, namely the inability to segment unknown objects (e.g., fallen debris from a moving truck, unusual moving obstacles, or rare animals) that lie outside the predefined label space. Recent vision language models \cite{hossain2025power,yuan2025sa2va,li2024womd} demonstrate strong zero-shot/few-shot recognition, grounding, and reasoning capabilities through text prompts. The recently released model, Sa2VA \cite{yuan2025sa2va}, combines SAM-2 with multimodal large language models (MLLMs) for dense visual grounding. However, these models still depend on linguistic descriptions, are biased to static per-frame information, have weaker ability to comprehend the temporal order~\cite{zohar2025apollo,xue2025seeing}, and require higher computational resources. Our method eliminates the need for language supervision, operating directly on motion patterns to detect unknown objects with lower computational overhead, making it more suitable for real-time applications. We believe in robotics applications, there will be an interplay between such general-purpose computationally intensive MLLMs and light-weight efficient techniques meant for reliability and safety, such as ours. 

AVOS methods operate without manual initialization and infer object masks directly from the raw video stream. Crucially, AVOS offers a foundation for video class-agnostic segmentation \cite{Siam_2021_CVPR}, a task that explicitly requires segmenting objects in the scene, regardless of their semantic class. By leveraging appearance, motion, and geometry cues from monocular video, this paradigm enables the detection of unknown objects. To overcome limitations of efficiency and the focus on only certain semantic classes, recent research has explored alternative strategies~\cite{inoue2024channel}. We focus on extending the video class agnostic segmentation to incorporate longer temporal context, where our video transformer takes an input clip of multiple RGB frames, i.e., more than two, and does not rely on explicit optical flow as input. 

Inspired by Mask2Former's multiscale transformer decoder, recent multiscale video transformer architectures~\cite{karim2023med, siam2023multiscale} proposed variants that are class and instance agnostic, without the use of per-segment mask classification loss. These methods can readily be extended to semantic segmentation and do not require instance, panoptic, or tracking annotations in their training. This is especially needed in domains where it is expensive to acquire such annotations, e.g., in the medical, robotics, or remote sensing domains. Inspired by these, our work focuses on refining the multiscale decoding mechanism to further improve its efficiency while maintaining strong performance. We propose a multi-stage multiscale query-memory transformer decoder that augments compressed queries with shared-memory attention and operates directly on high-resolution spatiotemporal feature maps. Additionally, we propose a computationally efficient mechanism using scale-specific random drop-token reducing its GPU memory footprint and inference time.

\section{Method}

\begin{figure}[t]
    \centering
    \includegraphics[width=\textwidth]{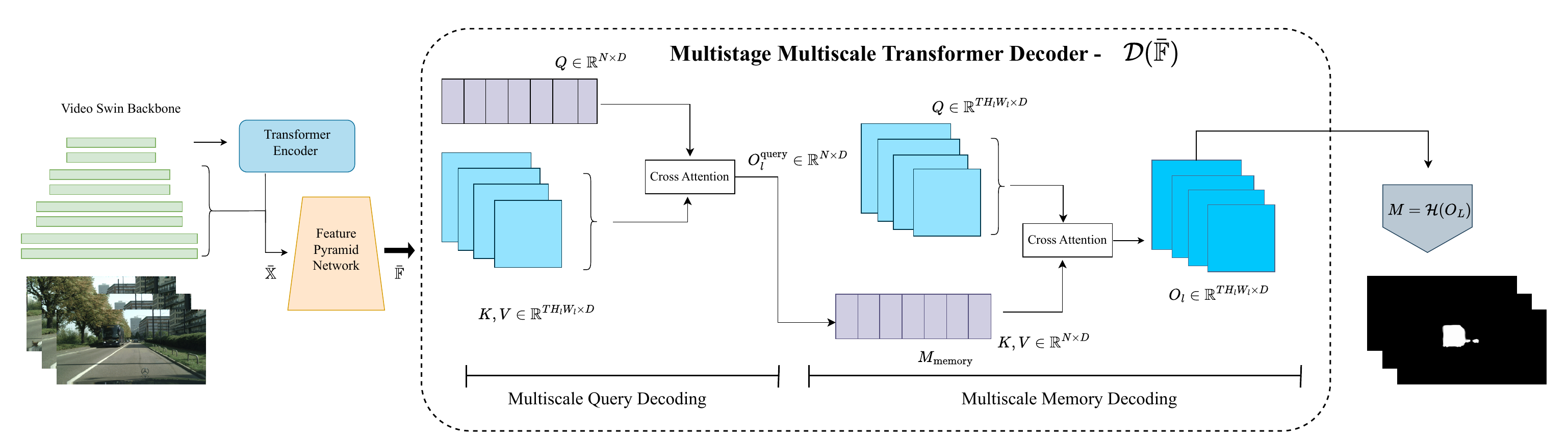}
    \caption{Overview of our proposed multi-stage multiscale query-memory transformer architecture for video class-agnostic segmentation in autonomous driving. It extracts spatiotemporal features at multiple scales, $\bar{\mathbb{X}}$, using a Video-Swin backbone, followed by a transformer encoder and the FPN, resulting in the pyramidal features, $\bar{\mathbb{F}}$. These are processed through our two-stage multiscale query-memory transformer decoder, $\mathcal{D}$. In the first stage, a multiscale query decoding where a set of compressed learned queries, $O_l^{\text{query}}$,  for each scale, $l$, is used to represent the different segments in the clip. This is followed by a second stage of multiscale memory decoding where they are used as a shared learnable memory to refine the spatiotemporal context per scale, $O_l$, while preserving the fine-grained information. Each stage includes an iterative multiscale exchange using cross attention between the queries, $Q$, and the keys, values, $K, V$, respectively. We show only the last exchange and how its output is used, and omit the self attention operation for simplicity. The final high-resolution features are decoded by a segmentation head to produce dense binary motion masks, $M= \mathcal{H}(O_L)$.}
    \label{fig:arch}
    \vspace{-1em}
\end{figure}

\textbf{Architecture Overview.} Our work proposes an end-to-end multiscale video transformer purposed for video class agnostic segmentation that can be used in an autonomous driving setting, without the use of optical flow input. Our architecture is summarized in Figure~\ref{fig:arch}, which takes as input a clip of $T$ frames with raw RGB images and outputs the respective class-agnostic segmentation masks. It includes, a convolutional backbone relying on Video-Swin~\cite{liu2022video} that extracts spatiotemporal features, $\mathbb{X} = \{X_l\}^{L}_{l=1}$, for $L$ scale levels. Each feature map \(X_l\) is then reduced and flattened, $\bar{\mathbb{X}} = \{ \bar{X}_l \}^{L}_{l=1}$, where, $\bar{X}_l \in \mathbb{R}^{T H_l W_l \times D}$, \(H_l\) and \(W_l\) are the spatial dimensions, and \(D\) is the feature dimension. The coarsest scale representations are passed through a transformer encoder that applies self-attention, and are fused with the multiscale representations through a feature pyramid network (FPN)\cite{lin2017feature}, $\bar{\mathbb{F}}=\mathcal{E}(\bar{\mathbb{X}}$). These features are fed into a multi-stage multiscale query-memory decoder transformer, $O_L = \mathcal{D}(\bar{\mathbb{F}})$, which preserves spatial information. The output, $O_L$, represents the dense features at the highest resolution after such multiscale exchange, which ensures better region-level and pixel-level interactions across the input spatiotemporal features. This is followed by a lightweight segmentation head $M = \mathcal{H}(O_L)$ that operates directly on these spatiotemporal features to produce dense binary masks  $M$ for each frame. The segmentation head encompasses multiple 3D convolutional layers.

\textbf{Multi-Stage Multiscale Query-Memory Transformer Decoder.} In this section, we detail the operation of our transformer decoder, $\mathcal{D}$. We typically utilize a compact representation in terms of learnable queries, $\mathcal{O}^{\text{query}}$, that can act as visual summaries of the main segments in the input image/clip. This operation is referred to as \textit{Multiscale Query Decoding}. Assuming a self-attention followed by cross-attention operations using multiheaded attention, $\mathcal{A}(Q, K, V)$, that takes input, $Q, K, V$ as the queries, keys and values respectively. We learn $N$ compact representations through cross-attention, $\mathcal{O}^{\text{query}}_l \in \mathbb{R}^{N \times D}$ for each scale, $l$, as follows,
\[
\mathcal{O}^{\text{query}}_l = \mathcal{A}(\mathcal{O}^{\text{query}}_{l-1} + \mathcal{P}, \bar{X}_l + \mathcal{P}^{\text{sc}}_l + \mathcal{P}^{\text{st}}_l, \bar{X}_l),
\]
where $\mathcal{P}, \mathcal{P}^{\text{sc}}_l, \mathcal{P}^{\text{st}}_l$ are the learnable query positional embeddings, learnable scale-level embeddings, and the fixed spatiotemporal positional embeddings, respectively. The cross attention operation takes the learnable queries from the previous scale, $l-1$, to learn the ones of the current scale, $l$. At the first cross attention we randomly initialize these queries during training. These representations have benefits in learning compact region-level information, which can improve the segmentation. However, they suffer from the loss of fine details and spatial precision during cross-scale exchange. Thus, we choose to perform a consecutive operation that utilizes these learned compact representations while retrieving the dense spatiotemporal details, which we refer to as \textit{Multiscale Memory Decoding}. We compute the attention between the input features, $\bar{X}_l$, at each scale level and a shared learnable memory $\mathcal{M}_{\text{memory}}$. Each output $\mathcal{O}_l$ is obtained by applying a cross attention operation that, similarly, incorporates positional embeddings, $\mathcal{P}, \mathcal{P}^{\text{sc}}_l, \mathcal{P}^{\text{st}}_l$, as described earlier,
\[
\mathcal{O}_l = \mathcal{A}\left( \bar{X}_l + \mathcal{P}^{\text{sc}}_l + \mathcal{P}^{\text{st}}_l, \mathcal{M}_{\text{memory}} + \mathcal{P}, \mathcal{M}_{\text{memory}} \right).
\]
In contrast to \textit{Multiscale Query Decoding}, the rich feature maps $\bar{X}_l$ act as queries and the shared learned memory $\mathcal{M}_{\text{memory}}$ acts as the keys and values. This design maintains the original spatial and temporal resolution, ensuring that the outputs $\mathcal{O}_l$ retain their dimensions $(T H_l W_l \times D)$, which is critical for dense prediction tasks. The shared learned memory, $\mathcal{M}_{\text{memory}}$, is initialized with the previously learned queries, $O^{\text{query}}_L$. Subsequently, the output $\mathcal{O}_l$ is fused with the next input $\bar{X}_{l+1}$ through a mixing operation,
\[
\bar{X}_{l+1} = \mathcal{M}(\bar{X}_{l+1}, \mathcal{O}_l) = \bar{X}_{l+1} + \mathcal{I}^{l}_{l+1}(\mathcal{O}_l),
\]
where $\mathcal{I}^{l}_{l+1}$ performs bilinear interpolation to match spatial resolutions from scale, $l$, to scale, $l+1$. This fusion allows for iterative refinement of features across scales. This \textit{multi-stage multiscale query-memory} is a variant of multiscale transformer decoding that improves the performance over the conventional single-stage multiscale query decoding. In each stage, the multiscale exchange across scales $\{1,2,..., L\}$ can be repeated multiple iterations, $R$, as a coarse-to-fine refinement. To reduce computation and memory usage, we adopt a random token-drop strategy in the decoder in the finest two scales specific to the self-attention operation, before the cross-attention is performed. This means fewer computations at the finest scales to ensure efficient memory foot-print and computations. A subset of tokens is randomly selected based on a \textit{token keep ratio} $r \in (0,1]$, e.g., at $r=0.5$ we only retain half the tokens. The refined finest-scale features, $O_L$, is used for prediction.

\section{Experimental Results}

\textbf{Experimental Setup.} We evaluate our method on DAVIS'16, and VCAS~\cite{Siam_2021_CVPR} dataset that includes two sections for KITTI and Cityscapes re-purposed for video class agnostic segmentation (i.e., VCAS-KITTI, VCAS-Cityscapes). All models are trained on clips with $T=5$ for 15 epochs, using $R=9$ transformer decoder layers and $L=4$ multiscale levels. Each transformer block has $D=384$ hidden dimensions and eight attention heads. We learn $N=5$ queries and consequently the number of memory entries. We use a learning rate of $1\times10^{-4}$ and a Video-Swin~\cite{liu2022video} pretrained backbone architecture. The token keep ratio is set to 0.5 through the main experiments. Training is carried out on an NVIDIA A100 GPU with 40 GB of memory. We compare two models: (i) \textit{Multiscale Query (Baseline)}, using the conventional multiscale video transformer decoder, and (ii) \textit{Multi-Stage Multiscale Query-Memory (Ours)}.

\noindent \textbf{Ablation Study.}\\
\noindent \textit{\textbf{Segmentation Accuracy.}} Our proposed video transformer consistently outperforms the multiscale baseline as shown in Table~\ref{tab:inference_results}. The results validate that memory-aware decoding preserves spatiotemporal detail more effectively than compressed queries. The two-stage decoding also promotes hierarchical refinement across scales, further improving segmentation accuracy in complex scenes. 

\begin{table}[t]
\setlength{\tabcolsep}{3pt}
\centering
\begin{tabular}{lccc}
\toprule
\textbf{Inference Method} & \textbf{DAVIS'16} & \textbf{VCAS-KITTI} & \textbf{VCAS-Cityscapes} \\
\midrule
Multiscale Query (Baseline)   & 81.3 & 52.7 & 56.3 \\
Multi-stage Multiscale Query-Memory (Ours)  & \textbf{84.2} & \textbf{54.6} & \textbf{59.3} \\
\bottomrule
\end{tabular}
\vspace{1em}
\caption{\textbf{Comparison to the baseline}, that uses conventional multiscale decoding across three datasets. Our method consistently improves over the baseline, which is an adapted version from Mask2Former~\cite{cheng2022masked}.}
\label{tab:inference_results}
\vspace{-1em}
\end{table}

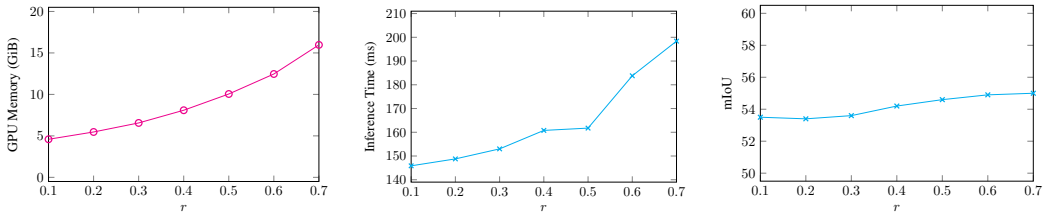
\begin{figure}[t]
    \centering

    \resizebox{0.32\textwidth}{!}{
    \begin{tikzpicture}
    \begin{axis} [
        width=\axisdefaultwidth,
        height=6cm,
        ymin=0, ymax=20,
        xmin=0.1, xmax=0.7,
        ytick = {0, 5, 10, 15, 20},
        xtick = {0.1, 0.2, 0.3, 0.4, 0.5, 0.6, 0.7},
        ylabel = GPU Memory (GiB),
        xlabel = $r$,
        enlarge x limits = {value = .1},
        enlarge y limits={abs=0.5},
    ]
    \addplot[magenta,mark=o,mark options={line width=0.9pt,scale=1.2,solid},style=solid] 
    coordinates {
        (0.1, 4.59) (0.2, 5.47) (0.3, 6.56) (0.4, 8.09) 
        (0.5, 10.05) (0.6, 12.46) (0.7, 15.97)
    };
    \end{axis}
    \end{tikzpicture}
    }%
    \hfill
    \resizebox{0.32\textwidth}{!}{
    \begin{tikzpicture}
    \begin{axis} [
        width=\axisdefaultwidth,
        height=6cm,
        ymin=140, ymax=210,
        xmin=0.1, xmax=0.7,
        ytick = {140, 150, 160, 170, 180, 190, 200, 210},
        xtick = {0.1, 0.2, 0.3, 0.4, 0.5, 0.6, 0.7},
        ylabel = Inference Time (ms),
        xlabel = $r$,
        enlarge x limits = {value = .1},
        enlarge y limits={abs=1.0},
    ]      
    \addplot[cyan,mark=x,mark options={line width=0.9pt,scale=1.2,solid},style=solid] 
    coordinates {
        (0.1, 145.87) (0.2, 148.78) (0.3, 153.00) 
        (0.4, 160.78) (0.5, 161.75) (0.6, 183.79) 
        (0.7, 198.38)
    };
    \end{axis}
    \end{tikzpicture}
    }%
    \hfill
    \resizebox{0.32\textwidth}{!}{
    \begin{tikzpicture}
    \begin{axis} [
        width=\axisdefaultwidth,
        height=6cm,
        ymin=50, ymax=60,
        xmin=0.1, xmax=0.7,
        ytick = {50, 52, 54, 56, 58, 60},
        xtick = {0.1, 0.2, 0.3, 0.4, 0.5, 0.6, 0.7},
        ylabel = mIoU,
        xlabel = $r$,
        enlarge x limits = {value = .1},
        enlarge y limits={abs=0.5},
    ]      
    \addplot[cyan,mark=x,mark options={line width=0.9pt,scale=1.2,solid},style=solid] 
    coordinates {
        (0.1, 53.5) (0.2, 53.4) (0.3, 53.6) 
        (0.4, 54.2) (0.5, 54.6) (0.6, 54.9) 
        (0.7, 55.0)
    };
    \end{axis}
    \end{tikzpicture}
    }

    \caption{\textbf{Token keep ratio, $r$, analysis} on the VCAS-KITTI dataset: (Left) GPU memory consumption in GB, (Middle) inference time per frame in milliseconds, and (Right) segmentation quality (mIoU).}
    \label{fig:kittimots_token_keep_all}
    \vspace{-1em}
\end{figure}

\textit{\textbf{Computational Efficiency.}} Figure~\ref{fig:kittimots_token_keep_all} shows the impact of varying the token keep ratio on GPU memory usage, inference time, and segmentation quality (mIoU) using the VCAS-KITTI data set. It only impacts the self-attention in our transformer decoder while using all the tokens in the cross-attention. As the token keep ratio increases from 0.1 to 0.7, GPU memory consumption increases significantly, indicating a higher computational cost with more retained tokens. Similarly, inference time per frame increases steadily. Interestingly, the mIoU curve shows a slight improvement with more tokens. Hence, using less tokens provide a clear gain in the computational efficiency metrics, with minimal degradation in the mIoU. Ratios more than 0.7 resulted in out of memory issues.

\section{Conclusion}

We present a multi-stage multiscale query-memory transformer decoder for video class-agnostic segmentation, targeting real-time safety in autonomous driving. Our memory-centric design preserves high-resolution spatiotemporal features while reducing computational overhead. Experimental results on DAVIS’16, KITTI, and Cityscapes confirm the superiority of our method over the conventional multiscale baseline. Our approach establishes an efficient transformer based foundation for unknown object detection in safety-critical robotics applications. We plan to explore further benchmarking efforts for such unknown objects segmentation, through carefully curating an evaluation benchmark outside the moving objects set available in the training datasets we use. Additionally, we plan to compare instance-wise vs. pixel-wise models, such as ours, with respect to their generalization and robustness on the aforementioned benchmark.


\end{document}